\documentclass{article}
\usepackage{spconf,amsmath,graphicx}
\usepackage{xcolor}
\usepackage{booktabs}
\usepackage{amsfonts}
\usepackage{adjustbox}

\newcommand{\yl}[1]{{\color{black}#1}} 

\title{SAMVG: A MULTI-STAGE IMAGE VECTORIZATION MODEL \\WITH THE SEGMENT-ANYTHING MODEL}
\name{Haokun Zhu$^{1, \sharp}$, Juang Ian Chong$^{1, \sharp}$, Teng Hu$^{1}$, Ran Yi$^{1,*}$, Yu-Kun Lai$^{2}$, Paul L. Rosin$^{2}$
\thanks{$^{\sharp}$ Both authors contributed equally to this research.}
\thanks{$^{*}$ Corresponding author.}
}
\address{$^1$Shanghai Jiao Tong University, $^2$Cardiff University\\
{\tt\small \{zhuhaokun,ianchong,hu-teng,ranyi\}@sjtu.edu.cn, \{LaiY4,RosinPL\}@cardiff.ac.uk}
}

\begin{document}
%
\maketitle
\begin{abstract}
Vector graphics are widely used in graphical designs and have received more and more attention. However, unlike raster images which can be easily obtained, acquiring high-quality vector graphics, typically through automatically converting from raster images, remains a significant challenge, especially for more complex images such as photos or artworks. In this paper, we propose SAMVG, a multi-stage model to vectorize raster images into SVG (Scalable Vector Graphics). Firstly, SAMVG uses general image segmentation provided by the Segment-Anything Model and uses a novel filtering method to identify the best dense segmentation map for the entire image. Secondly, SAMVG then identifies missing components and adds more detailed components to the SVG. Through a series of extensive experiments, we demonstrate that SAMVG can produce high quality SVGs in any domain while requiring less computation time and complexity compared to previous state-of-the-art methods.
\end{abstract}
\vspace{-10pt}
\section{Introduction}
\vspace{-7pt}
\label{sec:intro}
While raster images are commonly used for their adaptability and detailed representation, vector graphics have unique advantages. They represent images as mathematical equations rather than pixels, making them ideal for resizing without quality loss, which is especially useful for logos and icons.

Generating vector graphics, especially for complex images, is a challenging task compared to 
%
raster images \yl{which can be easily obtained}. Existing methods~\cite{sun2007image, orzan2008diffusion, xie2014hierarchical} 
generate \yl{vector graphics capable of faithfully reconstructing original images}.
%
\yl{But the produced representations tend to contain}
too many unnecessary parameters and fail to preserve essential topological features. Recent advances in deep learning and differentiable rendering have inspired new approaches~\cite{ma2022towards, li2020differentiable, carlier2020deepsvg, egiazarian2020deep, lopes2019learned, shen2021clipgen, reddy2021im2vec,hu2023stroke} aiming to create visually similar vector \yl{graphics} while also preserving topological features. However, deep generative models~\cite{carlier2020deepsvg, egiazarian2020deep, lopes2019learned, shen2021clipgen, reddy2021im2vec, su2021vectorization} face limitations related to high-quality training data availability and computationally intensive rendering-based training. Direct optimization algorithms~\cite{ma2022towards, li2020differentiable, yang2015effective} are more versatile but require good initialization for optimal results. The recent method LIVE~\cite{ma2022towards} addresses this by incrementally adding shape primitives, but can be computationally expensive for batch processing.
\begin{figure}[tbp]
    \centering
    \includegraphics[width=\linewidth]{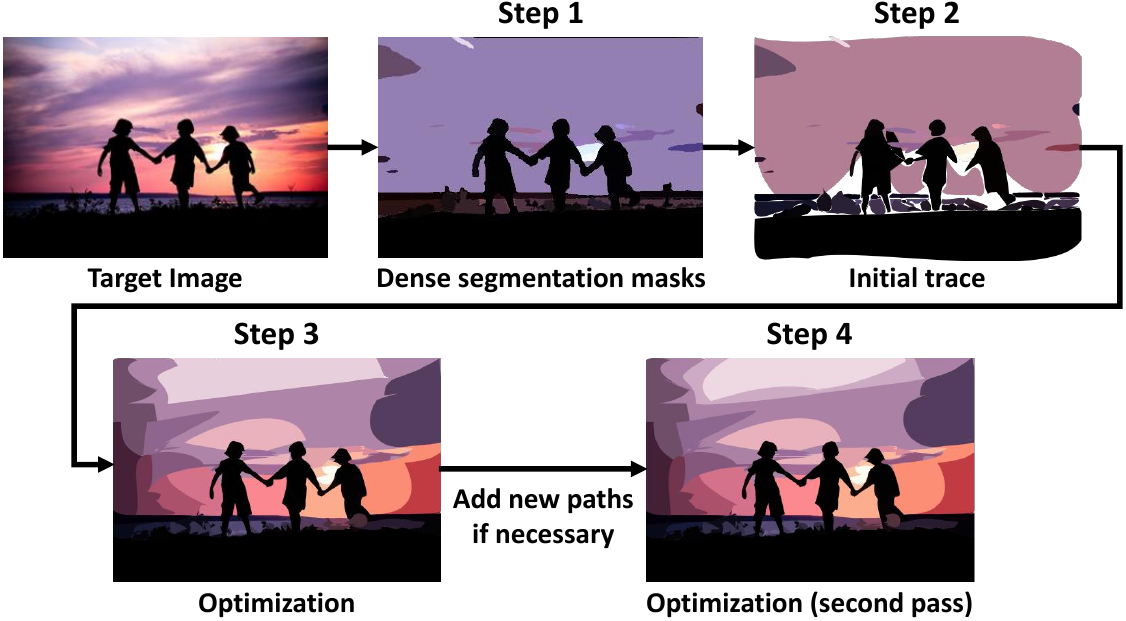}
    \vspace{-0.25in}
    \caption{High-level overview of SAMVG.}
    \label{fig:high-level framework}
    \vspace{-0.25in}
\end{figure}

In this paper, we aim to solve the initialization problem in direct optimization-based image vectorization methods in order to efficiently achieve high-quality results. A common image vectorization method involves segmenting images into their color components, as shown in previous research~\cite{diebel2008bayesian, selinger2003potrace}. However, as images become more complex, accurately identifying meaningful components becomes challenging. To address this problem, we propose SAMVG, a multi-stage image vectorization  model for converting raster images into high-quality SVGs with reasonable runtime. 

SAMVG utilizes Segment-Anything Model (SAM)~\cite{kirillov2023segment} to create segmentation masks for the entire image, which are then refined using a novel filter technique. Missed regions are addressed by instructing SAM to generate additional masks. These masks are used to trace each component into B\'ezier curves, forming an initial SVG. The SVG is further enhanced via differentiable rendering~\cite{li2020differentiable} to align with the target image. Lastly, regions needing more shape primitives are identified through error map convolution. Their centers are used as prompts for SAM to generate another set of masks, which are added to the SVG before a final round of optimization.

We assessed the effectiveness of SAMVG through experiments, testing its generalization ability across various domains. In our quantitative evaluation against state-of-the-art methods, SAMVG consistently outperformed previous approaches across multiple image evaluation metrics, simultaneously demonstrating significantly improved runtime efficiency. Furthermore, qualitative comparison between outputs of SAMVG and those of other methods highlights that SAMVG excels in generating shapes that better align with the semantic information of target images.

In summary, our contributions are three fold:



\begin{itemize}
    \vspace{-0.14in}
    \item We propose SAMVG, a multi-stage image vectorization  model for converting raster images into high-quality SVGs with reasonable runtime.
    \vspace{-0.10in}
    \item  We introduce a novel filter method to identify the best dense segmentation map for the entire image and propose to employ a novel convolution-based approach to identify regions requiring additional shape primitives for representation.
    \vspace{-0.10in}
    \item Extensive experiments show our superior performance over previous methods. Qualitative analysis further shows the ability of SAMVG to generate shapes that closely match the semantic content of target images.
\end{itemize}
\vspace{-17pt}
\section{Method}
\begin{figure}[htbp]
    \centering
    \includegraphics[width=\linewidth]{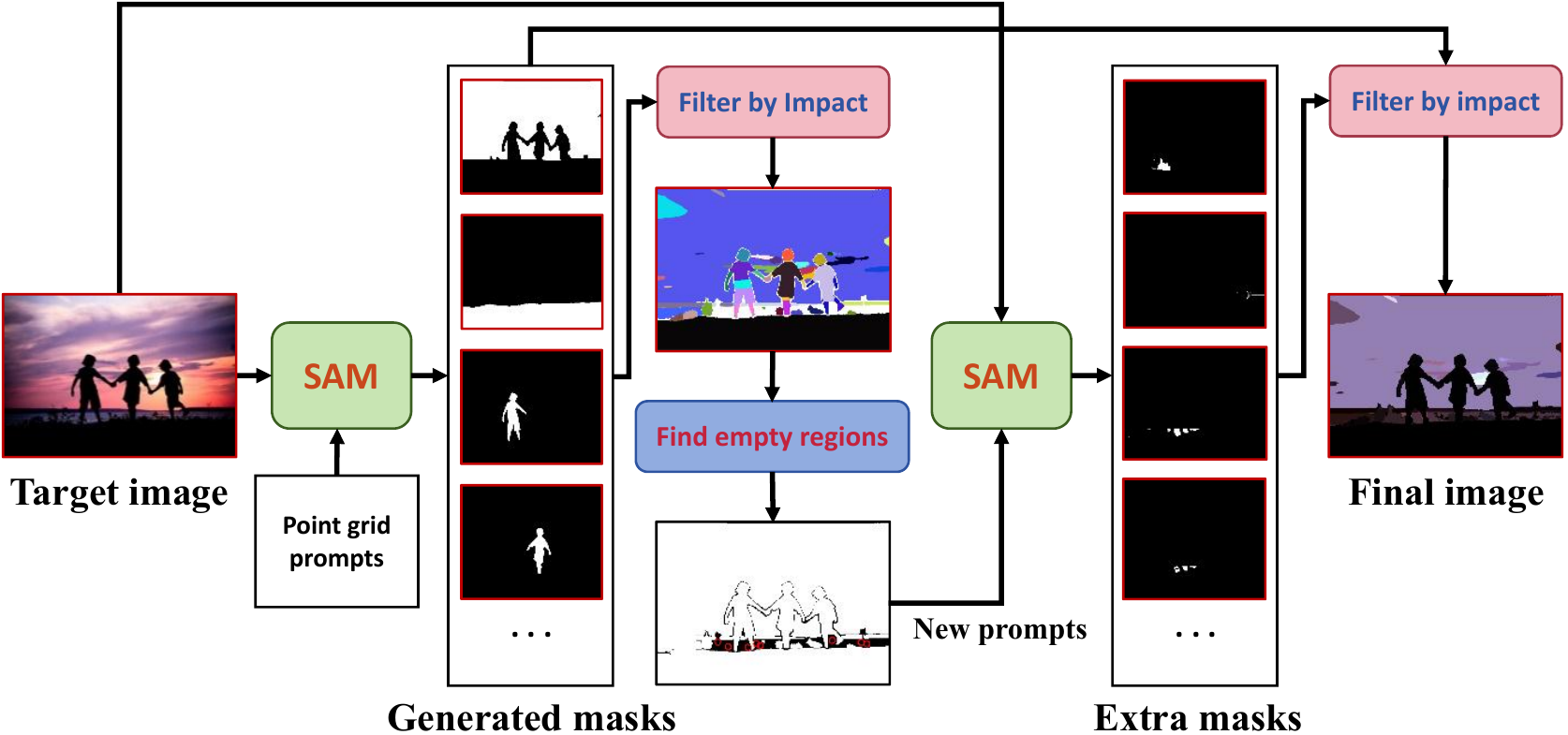}
    \vspace{-22pt}
    \caption{Flow diagram of the first stage of SAMVG to retrieve high quality segmentation masks. We filter the masks by testing its impact on the image render, and prompt the model twice for any components missed.}
    \label{fig:get_mask}
    \vspace{-13pt}
\end{figure}

\vspace{-8pt}
\label{sec:method}
SAMVG comprises four key stages: segmentation, filtering, tracing, and optimization, as shown in Fig.~\ref{fig:get_mask}. These stages are integral to the image vectorization process, and we provide a brief overview of each:

\textbf{(1) Retrieving Segmentation Masks.} Given a target image $I \in \mathbb{R}^{3 \times w \times h}$, SAM generates a list of segmentation masks $m_1,...,m_n\in\mathbb{R}^{w\times h}$. To boost segmentation quality, we locate the centers of missing components for a second round of prompts. We then filter out redundant masks and sort the remaining ones by area. \textbf{(2) Approximate Tracing.} For each component in the masks, the approximate shapes of the mask are traced with B\'ezier curves to produce an initial SVG denoted by $S$. \textbf{(3) Optimization.} We denote the rasterized image of $S$ as $I^{\prime} = R(S)$. $S$ is optimized through a specified number of iterations, utilizing mean square error (MSE) loss with respect to the target image $I$ and incorporating additional regularization losses to enhance shape smoothness and minimize artifacts. \textbf{(4) Identifying Missing Components.} We use convolution with a circular kernel on the difference map between $I^{\prime}$ and $I$ to detect any missing components. If such components are found, we add them to $S$ by repeating steps 1 and 2. Finally, the resulting SVG undergoes optimization to produce the ultimate result.

Fig.~\ref{fig:high-level framework} visually shows the intermediate results of each step in the SAMVG image vectorization process. Subsequent sections offer detailed explanations of each step, providing a comprehensive overview of the SAMVG and its components.
\vspace{-19pt}
\subsection{Stage 1: Retrieving Segmentation Masks}
\vspace{-3pt}
In the initial segmentation stage of SAMVG, we use the Automatic Masks Generator (AMG) from SAM. AMG is prompted with a customizable $32 \times 32$ grid of points, 
along with SAM's test-time augmentation stage in which multiple overlapping zoomed-in image crops
are included for comprehensive segmentation. 
It produces a list of masks, which are then filtered based on the confidence score and intersection over union (IoU). Post-processing removes small components and holes to aid path tracing in later stages. 

\begin{figure}[t]
    \centering
    \includegraphics[width=0.77\linewidth]{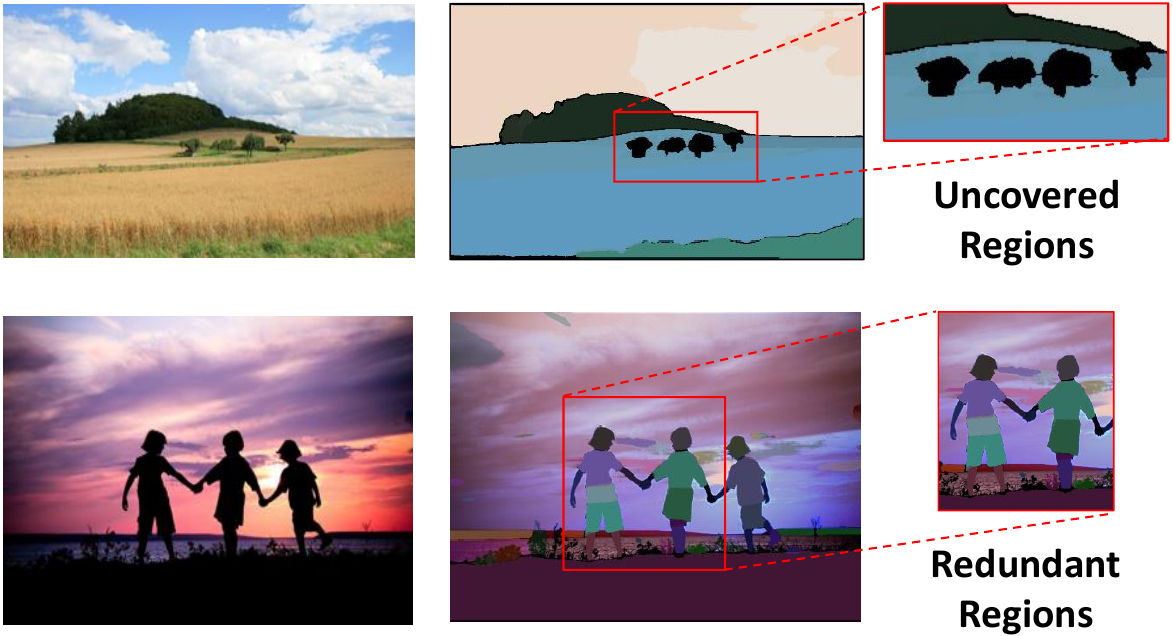}
    \vspace{-0.15in}
    \caption{Examples of failure cases from the Automatic Masks Generator provided in SAM.}
    \label{fig:failure case1}
    \vspace{-0.23in}
\end{figure}

The list of masks generated by AMG serves as a starting point, but it may not be sufficient for immediate tracing. Depending on the filter threshold and image complexity, AMG can predict either too few masks, leaving large areas uncovered or too many masks leading to redundancy (see Fig.~\ref{fig:failure case1}). To address this, we introduce a filtering method called ``Filter by Impact'' in addition to existing filters to ensure that all retained masks are both correct and significant. Additionally, we employ convolution with a circular kernel to identify potential center points in large uncovered regions, which are then used as prompts to generate additional segmentations.

In the following sub-sections, we provide detailed explanations of the filtering masks and the process of finding uncovered regions for prompts. 
\vspace{-0.15in}
\subsubsection{Filter by Impact}
\vspace{-0.08in}
To filter undesirable masks, we assess their impact by rendering them on a canvas. We begin with a blank canvas $C_0\in\mathbb{R}^{3 \times w\times h}$ and sort masks $m_1,...,m_n\in \{0, 1\}^{w\times h}$ by area in descending order. Each mask $m_i$ is assigned a color $c_i \in \mathbb{R}^3$ by averaging the regions it covers on $I$. We sequentially add mask $m_i$ to the canvas as follows:
\vspace{-5pt}
\begin{equation}
    C_{i}=f(C_{i-1},m_{i},c_{i}),
    \vspace{-0.05in}
\end{equation}
where $f\colon\mathbb{R}^{3 \times w\times h}\to\mathbb{R}^{3 \times w\times h}$ sets the color of pixels in $C_{i-1}$ covered by $m_i$ as $c_i$. Then we calculate the normalized mean square error. We ensure that pixels not covered by any masks on $C$ have the maximum error to prevent bias toward bright pixels. The impact $\gamma_i$ of $m_i$ is defined as the difference between the new and previous error:
\vspace{-0.07in}
\begin{equation}
\begin{array}{c}
e_{i}=\frac{||I-C_{i}||_{2}}{MaskArea}, \\
\gamma_i = e_i-e_{i-1}.
\end{array}
\vspace{-0.07in}
\end{equation}
If $\gamma_i$ falls below a predefined threshold, we discard $m_i$ and revert $C_i$ to the previous state $C_{i-1}$.

The ``Filter by Impact'' method in SAMVG offers several advantages. It retains masks representing sub-parts of objects with significant intersection over union (IoU) with parent object masks while filtering out small or incorrect masks. This ensures that all generated masks contribute significantly to the image representation. The filtering process is applied iteratively whenever SAM generates new masks, minimizing redundancy and enhancing vectorization efficiency by maintaining a list of relevant masks.

\vspace{-9pt}
\subsubsection{Locating Uncovered Regions}
\vspace{-4pt}
\label{subsec:Locating Uncovered Regions}
To identify large uncovered regions in $C$, we first generate an alpha mask $C_{\alpha}\in \mathbb{R}^{w \times h}=\vee _{i=1}^nm_i$, where $\vee$ stands for the bitwise-OR-operator. Convolution is then performed on $C_{\alpha}$ with a fixed circular kernel $k\in\{0,1\}^{r\times r}$ to get $C^\prime_{\alpha}$, where
\vspace{-0.04in}
\begin{equation}
k_{i, j}=\left\{\begin{array}{cl}
1, & \text { if } \sqrt{\left(i-\frac{r}{2}\right)^{2}+\left(j-\frac{r}{2}\right)^{2}} \leq r \\
0, & \text { otherwise }
\end{array} .\right.
\vspace{-0.04in}
\end{equation}
We identify candidate points by selecting the coordinates in $C^\prime_{\alpha}$ with zero values. Convolution identifies points belonging to large uncovered regions within a circle, whose radius is determined as a fraction of the image size. Subsequently, we employ mean shift clustering~\cite{cheng1995mean} on these candidate points to generate prompts for SAM. After filtering, we obtain a final list of masks $m_1,...,m_n$ ready for tracing.
\vspace{-8pt}
\subsection{Stage 2\&3: Approximate Tracing and Optimization}
\vspace{-3pt}
Our tracing algorithm is similar to~\cite{sarfraz2004automatic}. However, instead of choosing corner points based on local maxima, we select the global maximum as the first corner point. We then remove nearby points from potential corner candidates and search for the next corner point among the remaining candidates. This process is repeated until a predetermined number of corner points is reached. This modification aims to maintain a fixed number of segments in each path for simplicity and comparability with baselines that have a fixed segment count.
Following the tracing stage, we generate an initial SVG, providing an excellent starting point for optimization. Leveraging differentiable rendering~\cite{li2020differentiable}, we directly optimize the SVG parameters. The primary loss function employed is the MSE loss and LPIPS loss~\cite{zhang2018unreasonable} calculated between the rendered and target image. Additionally, we incorporate the Xing loss introduced by~\cite{ma2022towards}, which penalizes shapes prone to self-interaction and encourages topologically sound shapes.

The total loss function can be formulated as:
\vspace{-5pt}
\begin{equation}
    L = L_{MSE}+\lambda_{Xing} L_{Xing} + \lambda_{LPIPS} L_{LPIPS},
    \vspace{-5pt}
\end{equation}
where $\lambda_{MSE}, \lambda_{Xing}, \lambda_{LPIPS}$ are hyper-parameters.



\begin{table}[t]
\centering
\resizebox{1.0\linewidth}{!}{
\begin{tabular}{c|ccc}
\toprule
Metrics                                                                          & LIVE~\cite{ma2022towards}   & DIFFVG~\cite{li2020differentiable}          & SAMVG           \\ \hline
MSE$^{\times 10^{-3}}$ $\downarrow$                                              & 5.75    & 16.1            & \textbf{4.89}   \\ \hline
LPIPS $\downarrow $                                                              & 0.259   & 0.318           & \textbf{0.243}  \\ \hline
FID $\downarrow $                                                                & 188.54  & 209.25          & \textbf{184.24} \\ \hline
Complexity $\downarrow $                                                         & 11.53   & 12.17           & \textbf{11.32}  \\ \hline
Paths $\downarrow $                                                              & 59.96   & 59.96           & \textbf{57.54}  \\ \hline
Num of Parameters $\downarrow $                                                              & 2038    & 2038            & \textbf{1956}   \\ \hline
Time (s) $\downarrow $                                                           & 2609.00 & 139.57          & \textbf{139.13} \\ \bottomrule
\end{tabular}}
\vspace{-9pt}
\caption{Quantitative results on the self-collected dataset, Complexity is proposed by~\cite{machado1998computing} and is calculated on how well images can be compressed by JPEG~\cite{wallace1991jpeg} standard.}
\vspace{-10pt}
\label{tab:quantitative results}
\end{table}
\vspace{-10pt}
\subsection{Stage 4: Identifying Missing Components}
After the first phase of optimization, SAMVG may still miss some semantically significant components if their size is too small. These components may carry semantic meanings that are crucial for human perception of the image. For example, SAMVG occasionally fails to display eyes when vectorizing portrait images due to their small size. However, when prompted at the correct location, SAM can provide appropriate segmentations for these components.


To address these missed components at the end of the first optimization phase, we detect them by convolving the difference map. We compute the difference map $D$ by summing across color channels using the target image $I$ and the current render $I^{\prime}$. We then apply convolution to $D$ with a fixed circular kernel, as described in Sec~\ref{subsec:Locating Uncovered Regions}, to eliminate noise, resulting in $D_1$. Subsequently, we apply thresholding to $D_1$, ensuring that:\
\vspace{-0.04in}
\begin{equation}
    \boldsymbol{D}_{2_{x,y}}=\left\{\begin{matrix}1, \; \text{  if} \; \boldsymbol{D}_{1_{x,y}}\geq\omega\\0, \; \text{  if} \; \boldsymbol{D}_{1_{x,y}}<\omega\end{matrix}\right.,
    \vspace{-0.04in}
\end{equation}
where $\omega$ is the threshold value experimentally determined to be 0.784. Following this, we use the centers of components, whose value equals 1, in $D_2$ as prompts to SAM to retrieve the masks. Finally, the masks are filtered by impact with respect to $I^{\prime}$ as the starting canvas, then traced and optimized for another 500 iterations to achieve the final SVG.

\section{Experiment}
\label{sec:experiment}

\subsection{Experiment Setup}
\vspace{-0.05in}
We use the Adam optimizer for all experiments, conducted on an Nvidia GeForce RTX 3090 GPU. The learning rates are 0.01 for color parameters and 1 for point parameters in all algorithms. All algorithms run for 1000 iterations. To ensure fairness, we optimize the parameters for 1000 iterations for all methods.
We assess methods on a self-collected dataset, comprising 120 images from various categories.



\vspace{-15pt}
\subsection{Quantitative Results}
\begin{figure}[!t]
    \centering
    \includegraphics[width=0.83\linewidth]{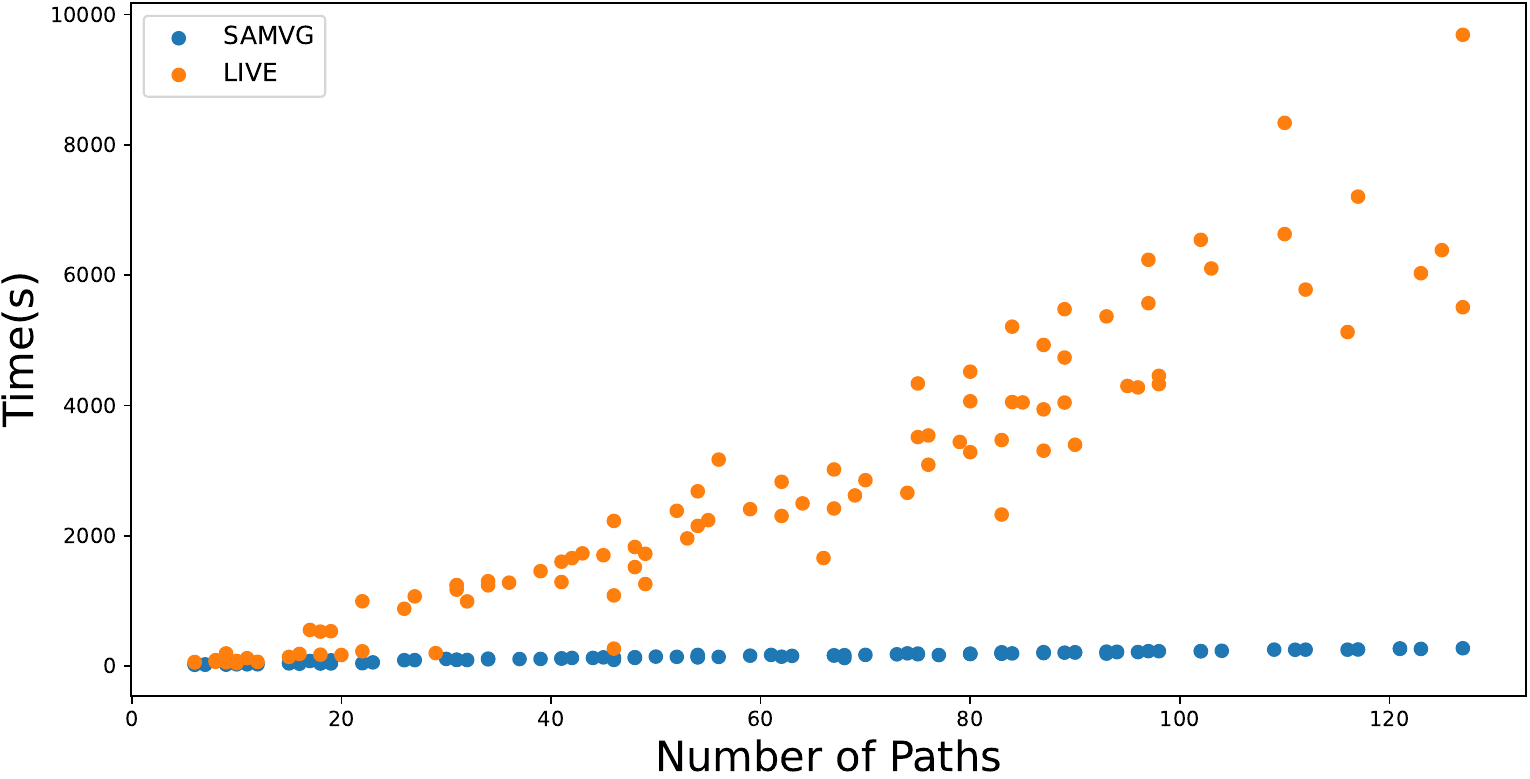}
    \vspace{-0.13in}
    \caption{Time taken by each algorithm vs. number of paths.}
    \label{fig:time}
\end{figure}

\vspace{-0.05in}
From Tab.~\ref{tab:quantitative results}, we can conclude that SAMVG outperforms LIVE and DIFFVG in all vectorization quality measures, including MSE, LPIPS, and FID. Besides, SAMVG is considerably faster than LIVE and offers significantly better vectorization quality than DIFFVG. Notably, the speed advantage of SAMVG becomes more pronounced for larger and more complex images, as demonstrated in Fig.~\ref{fig:time}, making it suitable for applications prioritizing fast and efficient vectorization.
\begin{table}[t]
\centering
\begin{tabular}{c|cc}
\toprule
Metrics                                                                          & SAMVG           & SAMVG w/o Filter \\ \hline
MSE$^{\times 10^{-3}}$ $\downarrow$                                              & 4.89            & \textbf{4.77}        \\ \hline
LPIPS $\downarrow $                                                              & 0.243  & \textbf{0.238}                \\ \hline
FID $\downarrow $                                                                & \textbf{184.24} & 230.64               \\ \hline
Complexity $\downarrow $                                                         & \textbf{11.32}  & 11.38                \\ \hline
Paths $\downarrow $                                                              & \textbf{57.54}  & 202.95               \\ \hline
Num of Parameters $\downarrow $  & \textbf{1956}   & 6696                 \\ \hline
Time (s) $\downarrow $                                                           & \textbf{142.00} & 563.96               \\ \bottomrule
\end{tabular}
\vspace{-0.10in}
\caption{Ablation Experiment on Filter by Impact.}
\vspace{-10pt}
\label{tab:ablation}
\end{table}

\vspace{-0.17in}
\subsection{Qualitative Analysis}
\vspace{-0.05in}
\begin{figure}[t]
    \centering
    \includegraphics[width=0.90\linewidth]{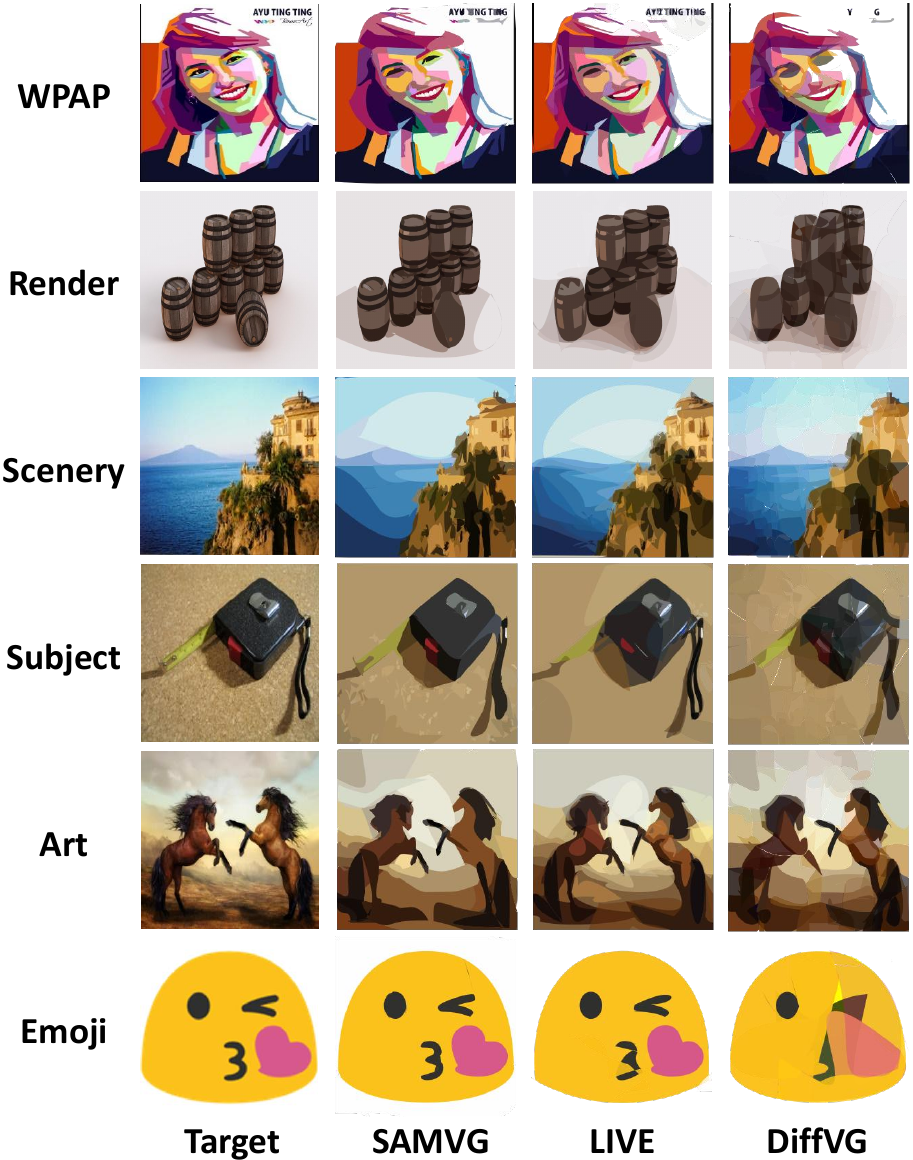}
    \vspace{-0.12in}
    \caption{Qualitative Comparisons between SAMVG, DIFFVG and LIVE.}
    \vspace{-0.22in}
    \label{fig:quality results}
\end{figure}
Fig.~\ref{fig:quality results} reveals that SAMVG generates vector graphics that closely resemble the target image. LIVE produces cleaner graphics with fewer extraneous shapes, but DiffVG generates patchy colors due to random initial path distribution. While the difference in vectorization quality between LIVE and SAMVG is subtle and subjective, SAMVG excels in terms of speed and the retrieval of semantic features from the target image. This ability to extract semantic features is a key advantage of SAMVG, enabling more accurate and meaningful image representations compared to other methods.

\subsection{Ablation Study}
To evaluate the efficacy of the proposed filter technique, \textbf{Filter by Impact}, we proceed to remove it from all stages of SAMVG and conduct an experiment using the same dataset. As observed in Table \ref{tab:ablation}, it becomes evident that the removal of Filter by Impact results in a substantial increase in path count, number of SVG parameters, and the processing time. Notably, this substantial change in metrics does not yield a commensurate enhancement in image quality, as discerned from the metrics reflecting image quality, namely MSE, LPIPS, FID and Complexity, which exhibit minimal change.
In light of these findings, we can infer that the proposed Filter by Impact method effectively improve the efficiency of the image vectorization process without compromising image quality.


\vspace{-0.15in}
\section{Conclusion}
\vspace{-10pt}
\label{sec:conclusion}
In this work, we present a novel multi-stage image vectorization model, SAMVG, which combines deep learning segmentation technology with traditional image vectorization. With the novel filter method, Filter by Impact, and a coarse-to-fine framework,  SAMVG exhibits the remarkable capability to produce SVGs of exceptional quality at a significantly enhanced rate. Through abundant experiments, we demonstrate that the vectorization quality of SAMVG is superior to the previous state-of-the-art methods while concurrently exhibiting enhanced operational efficiency.
\vspace{-10pt}
\section{Acknowledgement}
\vspace{-8pt}
This work was supported by National Natural Science Foundation of China (62302297, 72192821, 62272447), Young Elite Scientists Sponsorship Program by CAST (2022QNRC 001), Shanghai Sailing Program (22YF1420300), Beijing Natural Science Foundation (L222117), the Fundamental Research Funds for the Central Universities (YG2023QNB17), Shanghai Municipal Science and Technology Major Project (2021SHZDZX0102), Shanghai Science and Technology Commission  (21511101200).

\bibliographystyle{IEEEbib}
\bibliography{refs}

\begin{thebibliography}{10}

\bibitem{sun2007image}
Jian Sun, Lin Liang, Fang Wen, and Heung-Yeung Shum,
\newblock ``Image vectorization using optimized gradient meshes,''
\newblock {\em ACM Transactions on Graphics (TOG)}, vol. 26, no. 3, pp. 11--es, 2007.

\bibitem{orzan2008diffusion}
Alexandrina Orzan, Adrien Bousseau, Holger Winnem{\"o}ller, Pascal Barla, Jo{\"e}lle Thollot, and David Salesin,
\newblock ``Diffusion curves: a vector representation for smooth-shaded images,''
\newblock {\em ACM Transactions on Graphics (TOG)}, vol. 27, no. 3, pp. 1--8, 2008.

\bibitem{xie2014hierarchical}
Guofu Xie, Xin Sun, Xin Tong, and Derek Nowrouzezahrai,
\newblock ``Hierarchical diffusion curves for accurate automatic image vectorization,''
\newblock {\em ACM Transactions on Graphics (TOG)}, vol. 33, no. 6, pp. 1--11, 2014.

\bibitem{ma2022towards}
Xu~Ma, Yuqian Zhou, Xingqian Xu, Bin Sun, Valerii Filev, Nikita Orlov, Yun Fu, and Humphrey Shi,
\newblock ``Towards layer-wise image vectorization,''
\newblock in {\em Proceedings of the IEEE/CVF Conference on Computer Vision and Pattern Recognition}, 2022, pp. 16314--16323.

\bibitem{li2020differentiable}
Tzu-Mao Li, Michal Luk{\'a}{\v{c}}, Micha{\"e}l Gharbi, and Jonathan Ragan-Kelley,
\newblock ``Differentiable vector graphics rasterization for editing and learning,''
\newblock {\em ACM Transactions on Graphics (TOG)}, vol. 39, no. 6, pp. 1--15, 2020.

\bibitem{carlier2020deepsvg}
Alexandre Carlier, Martin Danelljan, Alexandre Alahi, and Radu Timofte,
\newblock ``{DeepSVG}: A hierarchical generative network for vector graphics animation,''
\newblock {\em Advances in Neural Information Processing Systems}, vol. 33, pp. 16351--16361, 2020.

\bibitem{egiazarian2020deep}
Vage Egiazarian, Oleg Voynov, Alexey Artemov, Denis Volkhonskiy, Aleksandr Safin, Maria Taktasheva, Denis Zorin, and Evgeny Burnaev,
\newblock ``Deep vectorization of technical drawings,''
\newblock in {\em European Conference on Computer Vision}. Springer, 2020, pp. 582--598.

\bibitem{lopes2019learned}
Raphael~Gontijo Lopes, David Ha, Douglas Eck, and Jonathon Shlens,
\newblock ``A learned representation for scalable vector graphics,''
\newblock in {\em Proceedings of the IEEE/CVF International Conference on Computer Vision}, 2019, pp. 7930--7939.

\bibitem{shen2021clipgen}
I-Chao Shen and Bing-Yu Chen,
\newblock ``{ClipGen}: A deep generative model for clipart vectorization and synthesis,''
\newblock {\em IEEE Transactions on Visualization and Computer Graphics}, vol. 28, no. 12, pp. 4211--4224, 2021.

\bibitem{reddy2021im2vec}
Pradyumna Reddy, Michael Gharbi, Michal Lukac, and Niloy~J Mitra,
\newblock ``{Im2Vec}: Synthesizing vector graphics without vector supervision,''
\newblock in {\em Proceedings of the IEEE/CVF Conference on Computer Vision and Pattern Recognition}, 2021, pp. 7342--7351.

\bibitem{hu2023stroke}
Teng Hu, Ran Yi, Haokun Zhu, Liang Liu, Jinlong Peng, Yabiao Wang, Chengjie Wang, and Lizhuang Ma,
\newblock ``Stroke-based neural painting and stylization with dynamically predicted painting region,''
\newblock {\em arXiv preprint arXiv:2309.03504}, 2023.

\bibitem{su2021vectorization}
Hao Su, Jianwei Niu, Xuefeng Liu, Jiahe Cui, and Ji~Wan,
\newblock ``Vectorization of raster manga by deep reinforcement learning,''
\newblock {\em arXiv preprint arXiv:2110.04830}, 2021.

\bibitem{yang2015effective}
Ming Yang, Hongyang Chao, Chi Zhang, Jun Guo, Lu~Yuan, and Jian Sun,
\newblock ``Effective clipart image vectorization through direct optimization of bezigons,''
\newblock {\em IEEE Transactions on Visualization and Computer Graphics}, vol. 22, no. 2, pp. 1063--1075, 2015.

\bibitem{diebel2008bayesian}
James~Richard Diebel,
\newblock {\em Bayesian Image Vectorization: the probabilistic inversion of vector image rasterization},
\newblock Ph.D. thesis, Stanford University, 2008.

\bibitem{selinger2003potrace}
Peter Selinger,
\newblock ``Potrace: a polygon-based tracing algorithm,'' 2003.

\bibitem{kirillov2023segment}
Alexander Kirillov, Eric Mintun, Nikhila Ravi, Hanzi Mao, Chloe Rolland, Laura Gustafson, Tete Xiao, Spencer Whitehead, Alexander~C Berg, Wan-Yen Lo, et~al.,
\newblock ``Segment anything,''
\newblock {\em arXiv preprint arXiv:2304.02643}, 2023.

\bibitem{cheng1995mean}
Yizong Cheng,
\newblock ``Mean shift, mode seeking, and clustering,''
\newblock {\em IEEE Transactions on Pattern Analysis and Machine Intelligence}, vol. 17, no. 8, pp. 790--799, 1995.

\bibitem{sarfraz2004automatic}
Muhammad Sarfraz and Murtaza Khan,
\newblock ``An automatic algorithm for approximating boundary of bitmap characters,''
\newblock {\em Future Generation Computer Systems}, vol. 20, no. 8, pp. 1327--1336, 2004.

\bibitem{zhang2018unreasonable}
Richard Zhang, Phillip Isola, Alexei~A Efros, Eli Shechtman, and Oliver Wang,
\newblock ``The unreasonable effectiveness of deep features as a perceptual metric,''
\newblock in {\em Proceedings of the IEEE/CVF Conference on Computer Vision and Pattern Recognition}, 2018, pp. 586--595.

\bibitem{machado1998computing}
Penousal Machado and Am{\'\i}lcar Cardoso,
\newblock ``Computing aesthetics,''
\newblock in {\em Brazilian Symposium on Artificial Intelligence}. Springer, 1998, pp. 219--228.

\bibitem{wallace1991jpeg}
Gregory~K Wallace,
\newblock ``The {JPEG} still picture compression standard,''
\newblock {\em Communications of the ACM}, vol. 34, no. 4, pp. 30--44, 1991.

\end{thebibliography}

\end{document}